\documentclass[sigconf]{acmart}

\usepackage{booktabs} 
\usepackage{epsfig}
\usepackage{graphicx}
\usepackage{amsmath}
\usepackage{amssymb}
\usepackage{multirow}
\usepackage{array}
\newcolumntype{H}{>{\setbox0=\hbox\bgroup}c<{\egroup}@{}}

\usepackage{ragged2e} 
\newcolumntype{P}[1]{>{\RaggedRight\hspace{0pt}}p{#1}}
\usepackage{etoolbox, xcolor, soul,tabularx, flushend,amsmath,graphicx,epstopdf,rotating,subcaption,amssymb}
\usepackage[normalem]{ulem}
\usepackage[algoruled,boxed,lined]{algorithm2e}

\definecolor{darkgreen}{RGB}{0,100,38}

\newcommand{\ie}{\emph{i.e.\;}}
\newcommand{\eg}{\emph{e.g.\;}}
\newcommand{\etal}{\emph{et al.\;}}

\newcommand{\cf}{\emph{c.f.\;}}

\newcommand{\att}[1]{\textcolor{black}{{#1}}} 

\newcommand*{\origrightarrow}{}

\renewcommand*{\textrightarrow}{\fontfamily{cmr}\selectfont\origrightarrow}

\iffalse
\setcopyright{acmcopyright}

\acmDOI{10.1145/3240508.3240599}
\acmISBN{978-1-4503-5665-7/18/10}

\acmYear{2018}
\copyrightyear{2018}
\acmConference[MM '18]{2018 ACM Multimedia Conference}{October 22--26, 2018}{Seoul, Republic of Korea}
\acmBooktitle{2018 ACM Multimedia Conference (MM '18), October 22--26, 2018, Seoul, Republic of Korea}
\acmPrice{15.00}

\else 
\setcopyright{none}
\renewcommand\footnotetextcopyrightpermission[1]{} 
\fi 
\begin{document}
\title{PHD-GIFs: Personalized Highlight Detection \\for Automatic GIF Creation}

\author{Ana Garc\'{i}a del Molino}
\authornote{Work done while at gifs.com}
\affiliation{\institution{School of Computer Science and Engineering,\\ Nanyang Technological University, Singapore}}
\email{ana002@e.ntu.edu.sg}

\author{Michael Gygli} 
\authornotemark[1]
\affiliation{\institution{Google Research, Zurich}
}
\email{gygli@google.com}
\renewcommand{\shortauthors}{Garcia del Molino and Gygli}
\fancyhead{}

\begin{abstract}
    Highlight detection models are typically trained to identify cues that make visual content appealing or interesting for the general public, with the objective of reducing a video to such moments.
    However, this ``interestingness'' of a video segment or image is subjective.
    Thus, such highlight models provide results of limited relevance for the individual user.
    On the other hand, training one model per user is inefficient and requires large amounts of personal information which is typically not available.
    To overcome these limitations, we present a global ranking model which can condition on a particular user's interests.
    Rather than training one model per user, our model is personalized via its inputs, which allows it to effectively adapt its predictions, given only a few user-specific examples.
    To train this model, we create a large-scale dataset of users and the GIFs they created, giving us an accurate indication of their interests.    
     Our experiments show that using the user history substantially improves the prediction accuracy. On a test set of \att{$850$} videos, our model improves the recall by \att{$8\%$} with respect to generic highlight detectors.
    Furthermore, our method proves more precise than the user-agnostic baselines even with only one single person-specific example.
\end{abstract}

%
%


\maketitle
\section{Introduction}
\begin{figure}[t]
	\centering
	\includegraphics[width=\linewidth]{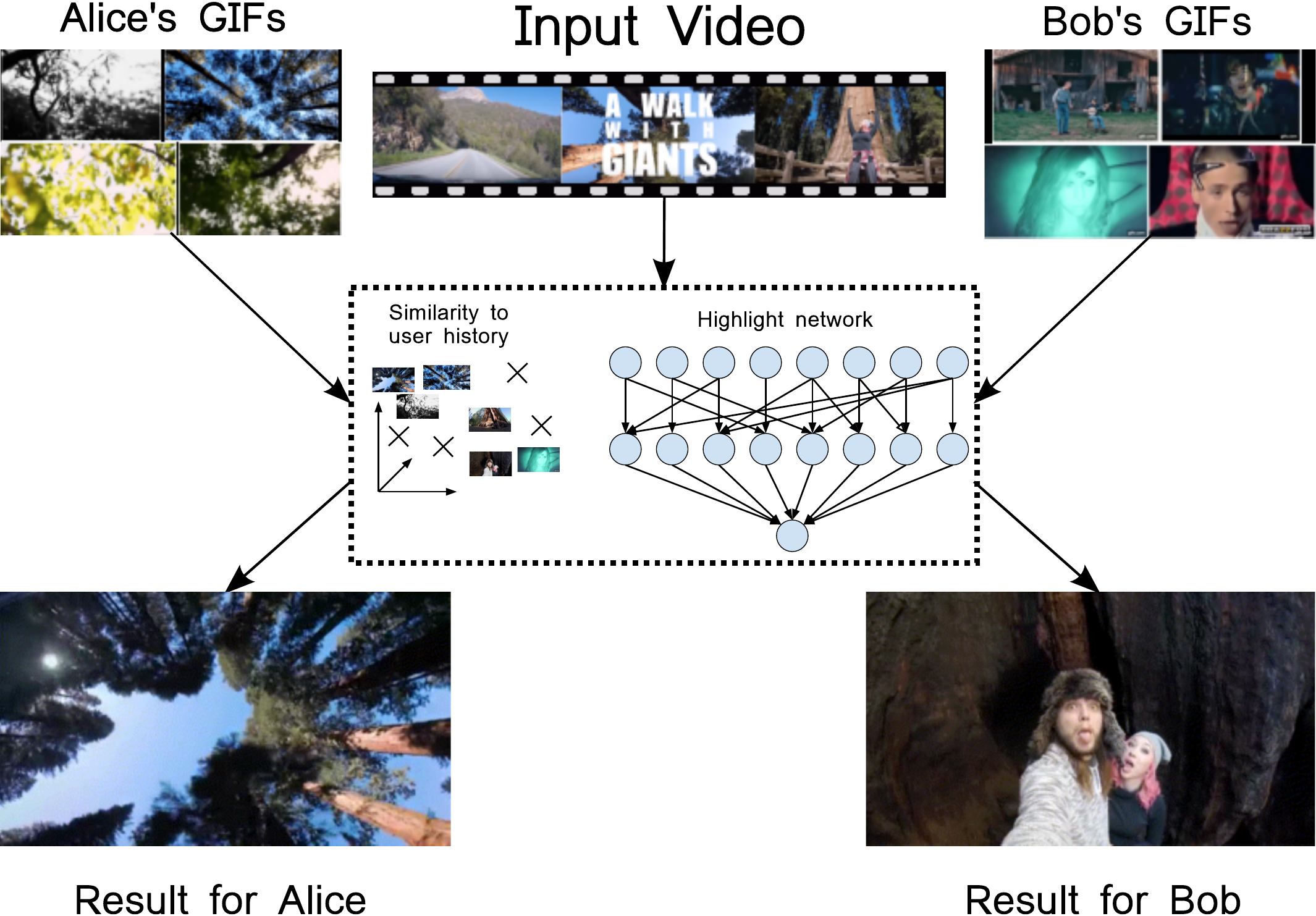}
	\caption{The notion of a highlight in a video is, to some extent, subjective. While previous methods trained generic highlight detection models, our method takes a user's previously selected highlights into account when making predictions. This allows to reduce the ambiguity of the task and results in more accurate predictions.}
    \label{fig:teaser}
\end{figure}

With the increasing availability of camera devices, more and more video is recorded and shared. In order to share these videos, however, they typically have to be edited to remove boring and redundant content and present the most interesting parts only. It's no coincidence that animated GIFs had a revival in the past years, as they make exactly this promise: that the video is reduced to the single most interesting moment~\cite{bakhshi2016fast}. 
Most users resort to online tools such as giphy, gifs.com or ezgif to create their GIFs manually, but video editing is usually a tedious and time consuming task. Recently, the research community has taken growing interest in automating the editing process~\cite{arev2014automatic,GygliSum15,gygli2016video2gif,
LeePred15,lin2015summarizing,PotapovSum,SunRank,yaohighlight,zhang2016summary,zhang2016video,ZhaoSum, yang2015unsupervised}.
These existing methods, however, share a common limitation, as they all learn a generic highlight detection or summarization model. This limits their potential performance, as not all users share the same interests~\cite{soleymani2015quest} and are thus editing video in different ways~\cite{AVS}.
One user may edit basketball videos to extract the slams, another one may just want to see the team's mascot jumping.
A third may prefer to see the kiss cam segments of the game.
An automatic method should therefore adapt its results to specific users, as exemplified in Figure~\ref{fig:teaser}.

\setcounter{footnote}{0}
In this work, we address the limitation of generic highlight detection and propose a model that explicitly takes a user's interests into account. 
Our model builds on the success of deep ranking models for highlight detection~\cite{gygli2016video2gif,yaohighlight}, but makes the crucial enhancement of making highlight detection personalized. Thereby, our model uses information on the GIFs a user previously created.
This allows the model to make accurate user-specific predictions, as a user's GIF history allows for a fine-grained understanding of their interests and thus provides a strong signal for personalization. This stands in contrast to relying on demographic data such as age or gender or interest in certain topics only. Knowing that a user is interested in basketball, for example, is not sufficient. A high-performing highlight detection model needs to have knowledge of what parts of basketball videos the user likes. Thus, to obtain a high-performing model, we need to collect information on a user's interest in certain objects or events~\cite{babaguchi2007learning, GygliInt13} and use that information for highlight detection.

To obtain that kind of data, we turn to gifs.com and its user base and collect a novel and large-scale dataset of users and the GIFs they created. On this data, we train several models for highlight detection that condition on the user history. 
Our experiments show that using the history allows making significantly more accurate predictions compared to generic highlight detection models, relatively improving upon the previous state of the art for automatic GIF creation~\cite{gygli2016video2gif} by \att{4.3\%} in MSD and \att{8\%} in recall.

To summarize, we make the following contributions:
\begin{itemize}
  \item A new large-scale dataset with personalized highlight information. It consists of $13,822$ users with $222,015$ annotations on $119,938$ videos. To the best of our knowledge, this is the first dataset with personalized highlight information as well as the biggest highlight detection dataset in general.
  We make the dataset publicly available\footnote{\url{https://github.com/gyglim/personalized-highlights-dataset}}.
  \item A novel model for personalized highlight detection (PHD). Our model's predictions are conditioned on a specific user by providing his previously chosen highlight segments as inputs to the model. This allows to 
  use all available annotations by training a single high-capacity model for all users jointly, while making personalized predictions at test time.
  \item An extensive experimental analysis and comparison to generic highlight detection approaches.
  \att{Our qualitative analysis finds that users often have high consistency in the content they select. Empirically, we show that our model can effectively use this history as a signal for highlight detection.  
  Our experiments further show the benefits of our approach over existing personalization methods: } our model improves over generic highlight detection, even when only one user specific example is available, and outperforms the baseline of training one model per user.
\end{itemize}


\section{Related Work}
Our work aims to predict what video segments a user is most interested in, using visual features alone.
It is thus a content-based recommender system~\cite{ricci2011introduction} for video segments, similar to
highlight detection and personalized video summarization. Our method further relates to collaborative filtering. In the following, we discuss the most relevant and recent works in these areas.
For an excellent overview and review of earlier video summarization and highlight detection techniques, we refer the reader to~\cite{truong2007video}.

\paragraph{Personalized video summarization.}
Early approaches in summarization cannot be personalized, as they are based on heuristics such as the occurrence of certain events,~\eg the scoring of a goal~\cite{truong2007video}. 
Exceptions are \cite{jaimes2002learning, agnihotri2005framework, babaguchi2007learning, takahashi2007user}, which build a user profile and use it for personalization.
Most notably, Jaimes~\etal~\cite{jaimes2002learning} also learn user-specific models directly from highlight annotation of a particular user.
All these methods, however, rely on annotated meta-data, rather than using only audio-visual inputs.

In the last years, methods using supervised learning on audio-visual inputs have become increasingly popular~\cite{ChuCosum, gong2014diverse, GygliSum15, LeeDisco12, ma2005generic, plummer2017enhancing, xu2015gaze, zhang2016summary, zhang2016video}.
These methods learn (parts of) a summarization model from annotated training examples.
Thus, they can be personalized by training on annotation coming from a single user, similar to~\cite{jaimes2002learning}. While that approach works in principle, it has two important practical issues. (i) Computational cost. Having a model per user is often infeasible in practice, due to the cost of training and storing models.
(ii) Limited data. Typically, only a small number of examples per user are available. This limits the class of possible methods to simple models that can be trained from a handful of examples. 
In contrast to that, we train a global model that is personalized via its inputs, by conditioning on the user history. This allows to train more complex models by learning from all users jointly. Thus, the proposed approach is able to perform well even for users that have not been seen in training and that have no examples to train with (\textit{cold start} problem).
Furthermore, as the user information is an input and is not embedded into the model parameters, our method does not need retraining as new user information arrives.

An alternative way to personalize summarization models is by analyzing the user behavior at recording~\cite{arev2014automatic, VariniPref} or visualization time~\cite{peng2011editing,zen2016mouse}, or requiring user input at inference time, either through specifying a text query~\cite{liu2015multi,sharghi2016query,vasudevan2017query,yang2003videoqa} or via an interactive approach~\cite{AVS, singla2016noisy}.
In the interactive approaches, the user gives feedback on individual proposals~\cite{AVS} or pairwise preferences~\cite{singla2016noisy}, which is then used to present a refined summary.
Instead, we do not require the user to know the full content of the video, nor require any input, such as user feedback, at test time: our model uses the user's history as the signal for personalization.

\paragraph{Highlight detection methods.}
The goal of highlight detection is to find the most interesting events of a video. In contrast to traditional video summarization approaches it does not aim to give an overview of the video, but rather just to extract the best moments~\cite{truong2007video}.
Recent methods for that task have used a ranking formulation, where the goal is to score interesting segments higher than non-interesting ones~\cite{gygli2016video2gif,SunRank, yaohighlight, jiao2017video,yu2018deep}. While~\cite{SunRank} used a ranking SVM model,~\cite{gygli2016video2gif,yaohighlight, jiao2017video,yu2018deep} trained a deep neural network using a ranking loss. Our work is similar to these approaches, in particular~\cite{gygli2016video2gif}, which also proposes a highlight detection model trained for GIF creation. But while they train a generic model, we use the user history to make personalized predictions.
\cite{soleymani2015quest} also predicts personalized interestingness, but does so on images and by training a separate model per user. It thus suffers from the same issues as~\cite{jaimes2002learning} and other existing supervised summarization methods that train one separate model per user. 
Ren~\etal~\cite{ren2017personalized} improve upon these methods by proposing a generic regression model which is personalized with a second, simpler model.
The second model predicts the residual of the generic model for a specific user. 
Thus, as our approach, this method can also handle users with no history, but it still requires (re-)training a model for each new user. 

\paragraph{Collaborative Filtering.}
In collaborative filtering (CF)~\cite{koren2009matrix}, interactions of users with items (\eg movie ratings) are used to learn user and item representations that accurately predict these and new interactions,~\eg a user's rating for a movie. CF has shown strong performance, for example in the Netflix challenge~\cite{bell2007lessons} and is used for video recommendation at YouTube~\cite{covington2016deep}.
While powerful, CF techniques cannot be easily applied to highlight detection, as that would require several interactions with the same video \textit{segment}. We find that in our data, few users create GIFs from the same video, let alone the same segment. This prevents learning a model from only interaction data alone.


\pagebreak
\section{Dataset}
\label{sec:dataset}
\begin{figure}[t]
\centering
\begin{tabular}[width=\linewidth]{@{}c@{}@{}c@{}}
\includegraphics[width=.5\linewidth]{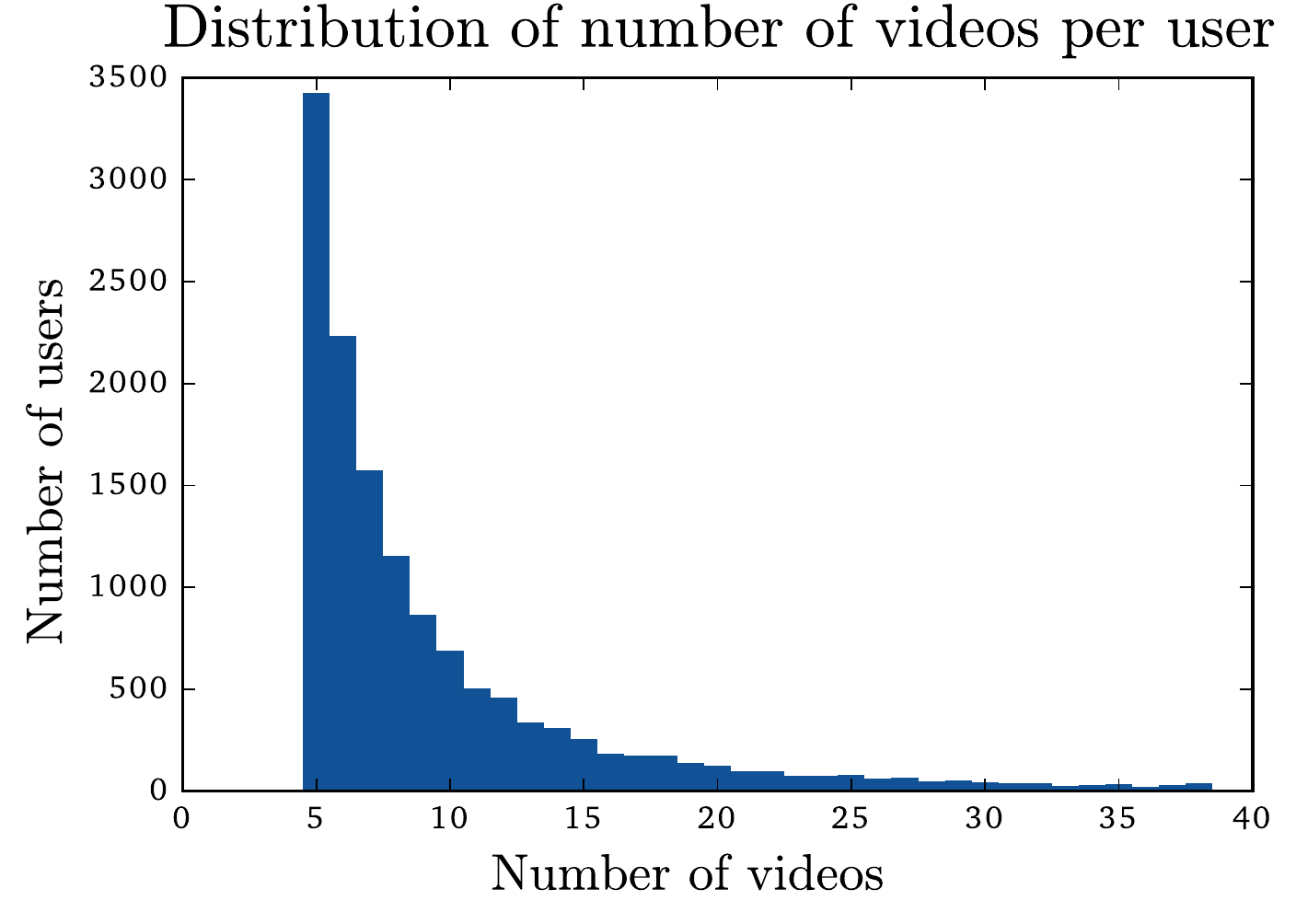} &\includegraphics[width=.5\linewidth]{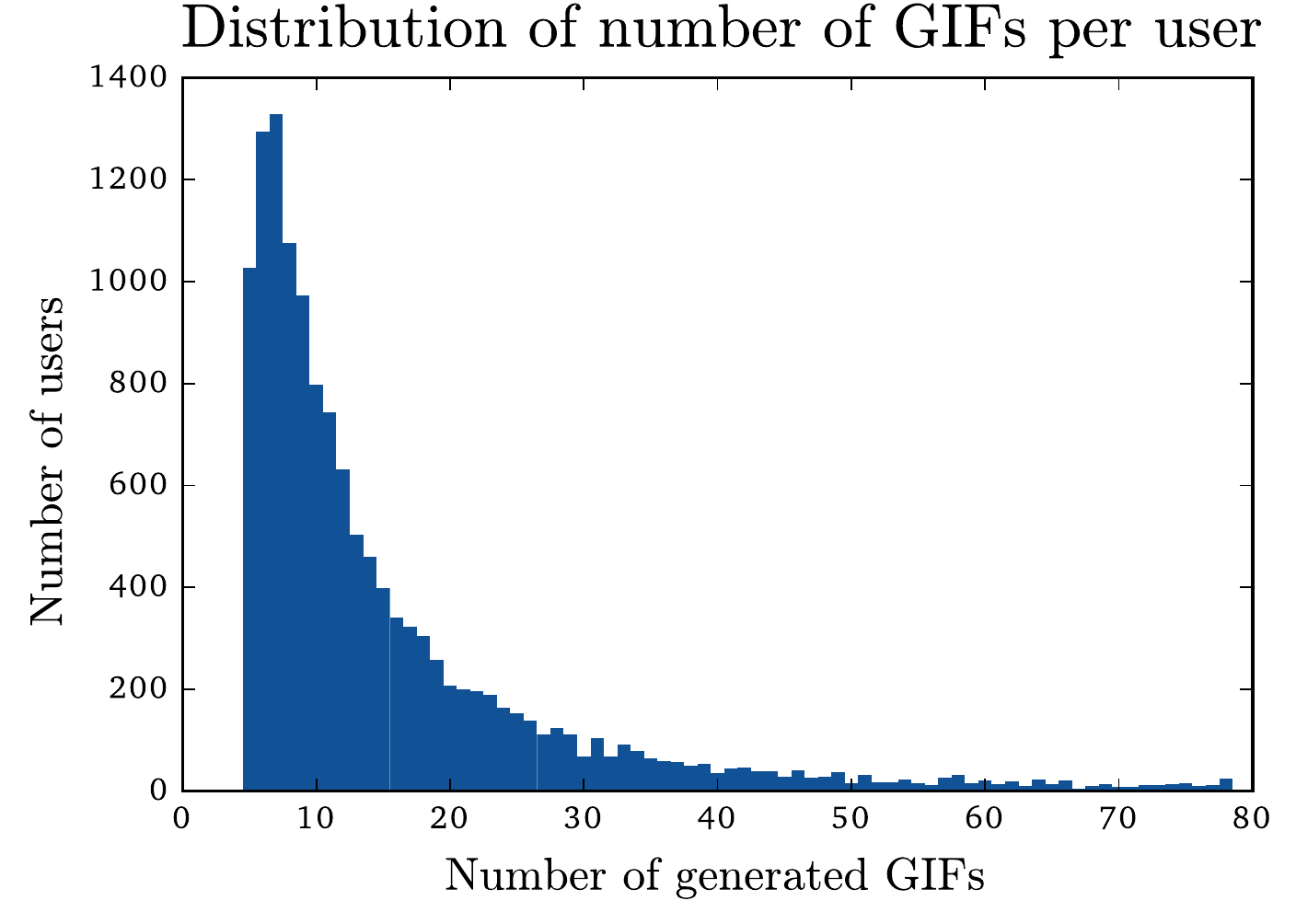} \\
    (a) & (b)\\
    \end{tabular}
	\vspace*{-7pt}
	\caption{The dataset in numbers: distribution of the amount of (a) videos per user, and (b) gifs per user.
    }
    \label{fig:DBstats}
\end{figure}

In order to be able to do personalized highlight detection, one key challenge is obtaining a training set that provides useful user information. Thereby, different kinds of user information is possible,~\eg meta-data on the user's age, gender, geographic location, what web editor was used and so on.
For our dataset, instead, we directly collect information on what video segments a specific user considers a highlight. Having this kind of data allows for strong personalization models, as specific examples of what a user is interested in help the model obtain a fine-grained understanding of that specific user. This stands in contrast to knowing demographic data, which would only allow to customize models based on loose indicators of interest such as the gender or location. Our idea of using a web video editor as a data source is similar to~\cite{gygli2016video2gif,SunRank}, but we additionally associate each GIF with a specific user, which allows for personalization.

\subsection{Data source}
To obtain personalized highlight data, we have turned to \emph{gifs.com} and its user base.
\emph{Gifs.com} is a video editor for the web and has a large base of registered users.
When a user creates a GIF,~\eg by extracting a key moment from a YouTube video, that GIF is linked to the user.
This allows to \att{query for} 
user profiles for users which have created several GIFs,~\ie contain a history that describes the user's interest. To have a reasonably sized sample of the users of interest, we restricted the selection to users that created GIFs from a minimum of five videos, where the last video is used for prediction, while the remaining ones serve as the history.
Thus, in our dataset, each user has a history of at least four videos.


\begin{figure}
	\centering
    \begin{subfigure}[b]{1\linewidth}
            \centering
	\includegraphics[width=.195\linewidth]{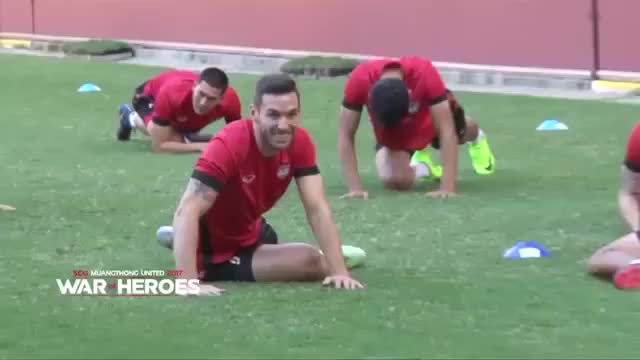}\,%
	\includegraphics[width=.195\linewidth]{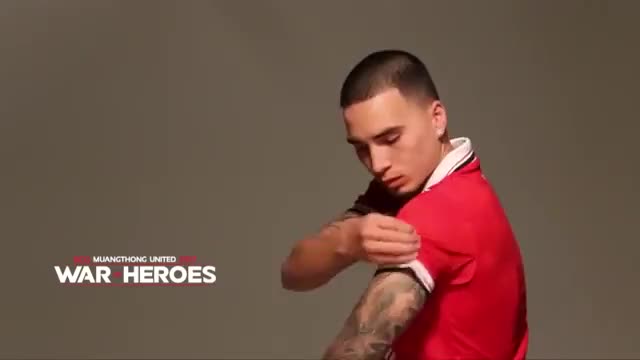}\,%
	\includegraphics[width=.195\linewidth]{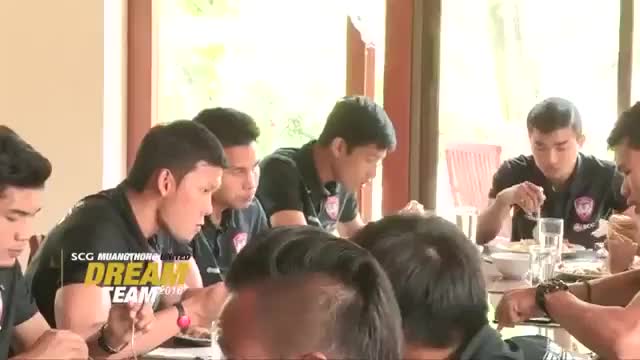}\,%
	\includegraphics[width=.195\linewidth]{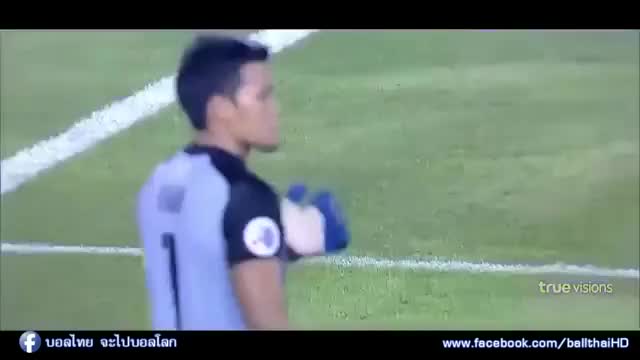}\,%
	\includegraphics[width=.195\linewidth]{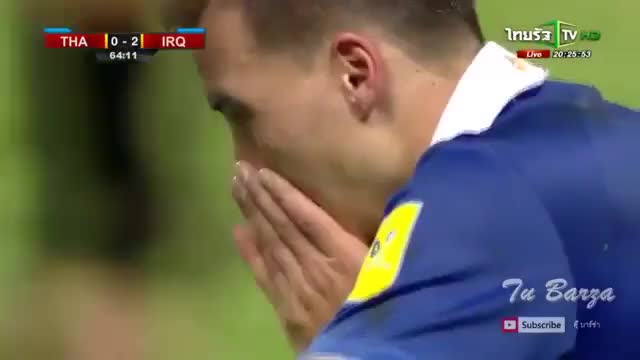}\,%
	\caption{Examples from a user consistently selecting GIFs of soccer players (202 GIFs). His interests differ from the majority of users, which consider goal scenes the most interesting~\cite{truong2007video}.}
    \end{subfigure}		
    
    \begin{subfigure}[b]{1\linewidth}
            \centering
	\includegraphics[width=.195\linewidth]{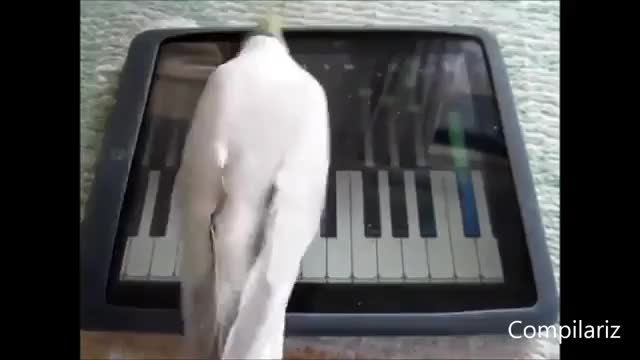}\,%
	\includegraphics[width=.195\linewidth]{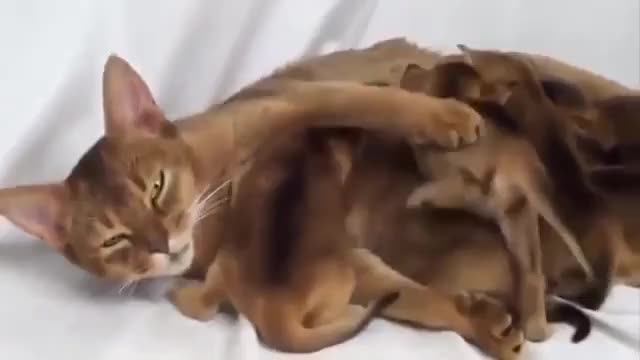}\,%
	\includegraphics[width=.195\linewidth]{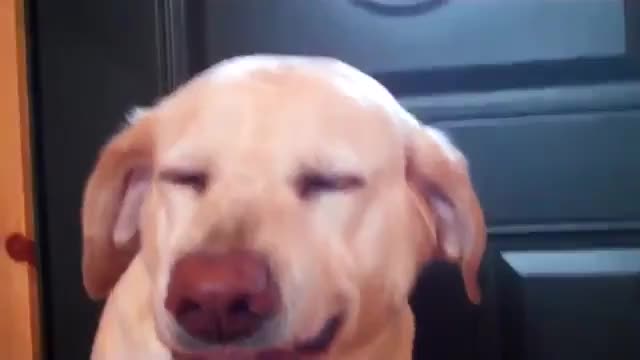}\,%
	\includegraphics[width=.195\linewidth]{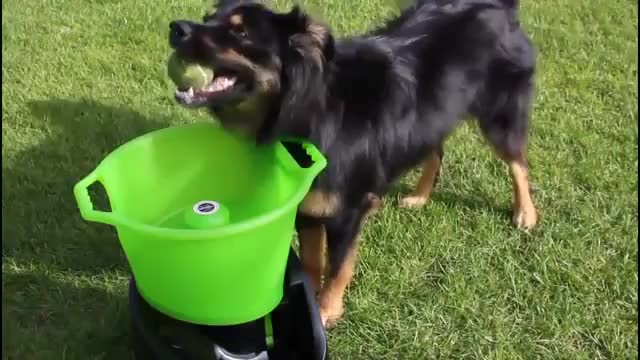}\,%
	\includegraphics[width=.195\linewidth]{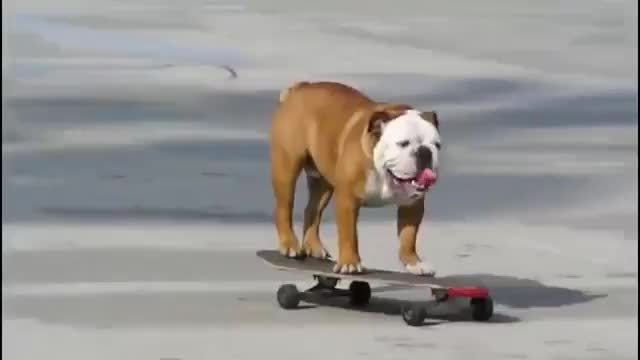}\,%
	\caption{Examples from a user consistently selecting GIFs of funny or cute pets  \att{(446 GIFs)}}
    \end{subfigure}		    
    \begin{subfigure}[b]{1\linewidth}
            \centering
\includegraphics[width=.195\linewidth]{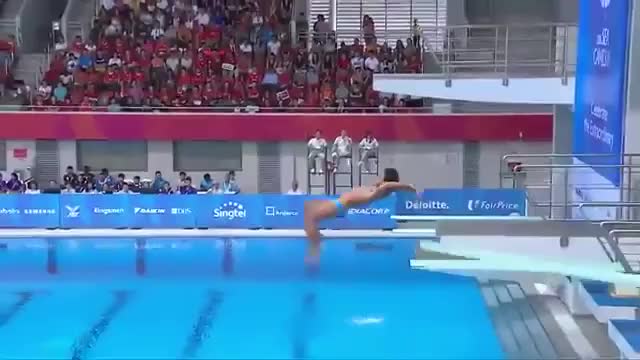}\,%
\includegraphics[width=.195\linewidth]{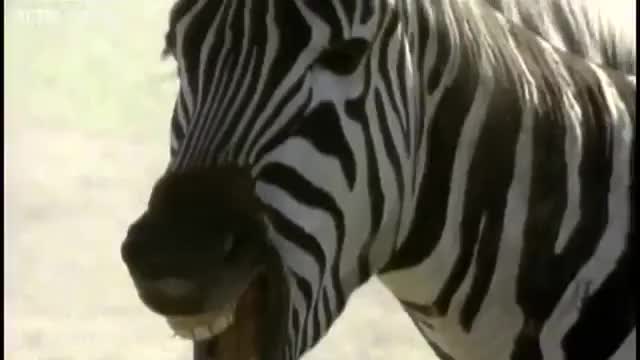}\,%
\includegraphics[width=.195\linewidth]{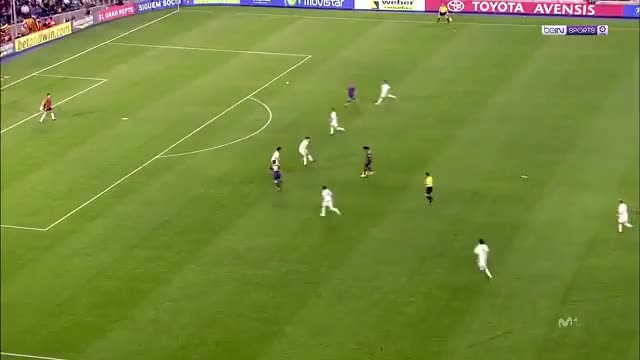}\,%
\includegraphics[width=.195\linewidth]{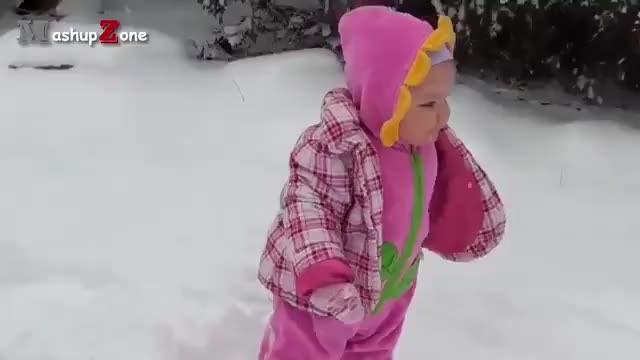}\,%
\includegraphics[width=.195\linewidth]{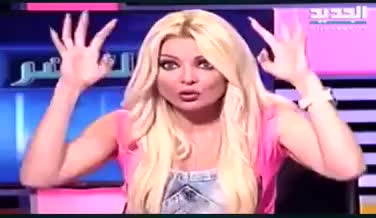}\,%
	\caption{Examples from a user with GIFs with interests in several categories like sports, funny animals and people \att{(21 GIFs)}.}
	\label{fig:multi_interest}
    \end{subfigure}
  	\vspace{-1.8\baselineskip}
	\caption{Example user histories (subsampled)}
    \label{fig:user_histories}
	\vspace{-.7\baselineskip}
\end{figure}

\begin{figure}[t]
	\centering
	\includegraphics[width=1\linewidth]{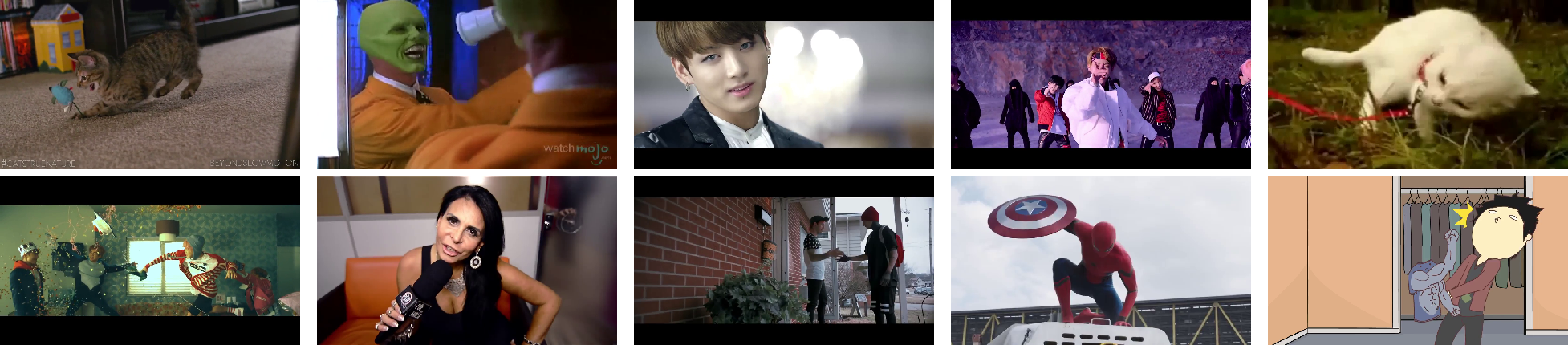}
    \vspace{-1.5\baselineskip}
	\caption{\textbf{Ten most selected moments.} The most popular moments in our dataset often show cats, music videos such as k-pop or famous movie scenes.}
    \label{fig:top_moments}
	\vspace{-1\baselineskip}
\end{figure}

\subsection{Analysis}
\label{sec:data_analysis}
Almost $14,000$ users on \emph{gifs.com} fulfilled our conditions. In total, the dataset contains $222,015$ annotations on $119,938$ YouTube videos. This is a significant leap with respect to other popular datasets such as the YouTube video highlight dataset~\cite{SunRank}, which contains about $4,300$ annotations, and the Video2GIF dataset~\cite{gygli2016video2gif}, which includes $100,000$ annotation in the form of GIFs.

Out of the \att{$14,000$} users, \att{$850$} were selected to form the test set (more details on the use of the dataset is given in Section~\ref{sec:implementation}). The selection was done such that the test videos (\ie the last video the user created a GIF from) are between $15$ seconds and $15$ minutes long, to avoid too simple scenarios as well as prevent extremely sparse labels suffering from chronological bias~\cite{song2015tvsum}. 
The distributions for the number of videos and GIFs per user in the full dataset are shown in Figure~\ref{fig:DBstats}.
Note that a user may generate more than one GIF from the same video, and thus the total amount of GIFs is greater than the number of videos.

In Figure~\ref{fig:user_histories} we show examples of users histories. When analyzing users we find that most have a clear focus,~\eg mostly or even exclusively create GIFs of funny pets. In some cases, users also have multiple interests (\cf Figure~\ref{fig:multi_interest}) and some have a clear focus with one or two outliers that show a different type of content. On the other hand, the most popular moments (most selected video segments in our dataset) show higher diversity. Their contents range from scenes with pets to interviews, cartoons, music videos and scenes of famous movies (see Figure~\ref{fig:top_moments}).
Given the high diversity of the dataset, and the consistent interests of specific users, 
we hypothesize that the user's history provides a reliable signal for predicting what GIFs he or she will create in the future.


\section{Method}
\label{sec:method}
In the following, we introduce our approach for highlight detection, which uses information about a user when making predictions.
In particular, we propose a model that predicts the score of a segment as a function of
both the segment itself and the user's previously selected highlights.
As such, the model learns to take into account the \textit{user history} to make accurate personalized predictions.

We define $V$ as the video from which a user $U$ wants to generate a GIF, and $s$ the segments that form it.
For our method, we use a ranking approach~\cite{joachims2002optimizing}, where a model is trained to score \att{positive video segments, $s^+$, higher than negative segments, $s^-$, from the same video.}
Thereby a segment is a positive if it was part of the user's GIF and a negative otherwise, as in~\cite{gygli2016video2gif}.
In contrast to previous works~\cite{SunRank,gygli2016video2gif,yaohighlight}, however, we do not make the predictions based on the segment alone, but also take a user's previously chosen highlights, their history, into account. Thus, our objective is 
\begin{equation}
\label{eq:objective}
h(s^+, \mathcal{G}) > h(s^-, \mathcal{G}), \:\:\:\:
\forall \left( s^+,s^- \right) \in V,
\end{equation}
\att{where $s^+$, $s^-$ are positive and negative segments coming from the same video $V$} and $h(s,\mathcal{G})$ is the score assigned to segment $s$.
$\mathcal{G}$ denotes all the GIFs that user $U$ previously generated,~\ie the \textit{user's history}.
Our formulation thus allows the model to personalize its predictions by conditioning on the user's previously selected highlights.

While there are several ways to do personalization, making the user history an input to the model has the advantage that a single model is sufficient and that the model can use all annotations from all users in training. A single model can predict personalized highlights for all users and new user information can trivially be included. Previous methods instead embedded the personal preferences into the model weights~\cite{soleymani2015quest,ren2017personalized}, which requires training one model per user and retraining to accommodate the new information. 

We propose two models for $h(\cdot, \cdot)$, which are combined with late fusion. One takes the segment representation and aggregated history as input (\textbf{PHD-CA}), while the second uses the distances between the segments and the history (\textbf{SVM-D}). Next, we discuss these two architectures in more detail. In all models we represent the segments $s$ and the history elements $g_i \in \mathcal{G}$ using C3D~\cite{tran2015learning} (conv5 layer). We denote these vector representations 
$\mathbf{s}$ and $\mathbf{g_i}$, respectively.

\subsection{Model with aggregated history}
\label{sec:FNN}
We propose to use a feed-forward neural network (FNN) similar to~\cite{gygli2016video2gif,yaohighlight}, but with the history as an additional input. More specifically, we average the history representations $\mathbf{g_i}$ across examples to obtain $\mathbf{p}$. The segment representation $\mathbf{s}$ and the aggregated history $\mathbf{p}$ are then concatenated and used as input to the model:
\begin{equation}
\label{eq:FNN}
h_{FNN}(s, \mathcal{G}) = FNN\left(\begin{bmatrix}
           \mathbf{s} \\
           \mathbf{p}
         \end{bmatrix}
         \right).
\end{equation}

As a model, we used a small neural network with 2 hidden layers with 512 and 64 neurons%
\footnote{
While different aggregation methods are possible, we found averaging the history to work well in practice.
We also tried alternative ways to aggregate, such as learning the aggregation with a sequence model (LSTM), but found this to lead to inferior performance (see Section~\ref{sec:experiments}).
}.

\subsection{Distance-based model}
\label{sec:distance-svm}
The assumption behind using a model of the form $h(s,\mathcal{G})$ is that the score of a segment depends on the 	similarity of the segment to a user's history. Thus, we investigated explicitly encoding that assumption into the model. Specifically, we create a feature vector that contains the cosine distances to the $k$ most similar history elements $g_i$. We denote this feature vector $\mathbf{d}$.
Using this representation we train a linear ranking model (ranking SVM~\cite{lee2014large}) to predict the score of a segment, \ie
\begin{equation}
\label{eq:svm}
h_{SVM}(s, \mathcal{G}) = \mathbf{w}^{\mathbf{T}}\mathbf{d} + b,
\end{equation}
where $\mathbf{w}$, $b$ are the learned weights and bias.
While the distance features could directly be provided to the model introduced in Section \ref{sec:FNN}, we find that training two separate models and combining them with late fusion leads to improved performance (\cf Table~\ref{tab:results_detail}). This is in line with previous approaches that found this method to be superior over fusing different modalities in a single neural network~\cite{simonyan2014two, carreira2017quo}.

\subsection{Model fusion}
\label{sec:LF}
We propose to combine the models introduced in Section \ref{sec:FNN} and \ref{sec:distance-svm} with late fusion.
As the models differ in the range of their predictions and their performance, we apply a weight for the model ensemble. 
To be concrete, the final prediction is computed as 
\begin{equation}
\label{eq:LF}
h(s, \mathcal{G}) = h_{FNN}(s, \mathcal{G}) + \omega * h_{SVM}(s, \mathcal{G}),
\end{equation} 
where $\omega$ is learned with a ranking SVM on the videos of a held out validation set.


\begin{figure}[t]
  \centering
  \includegraphics[width=\linewidth]{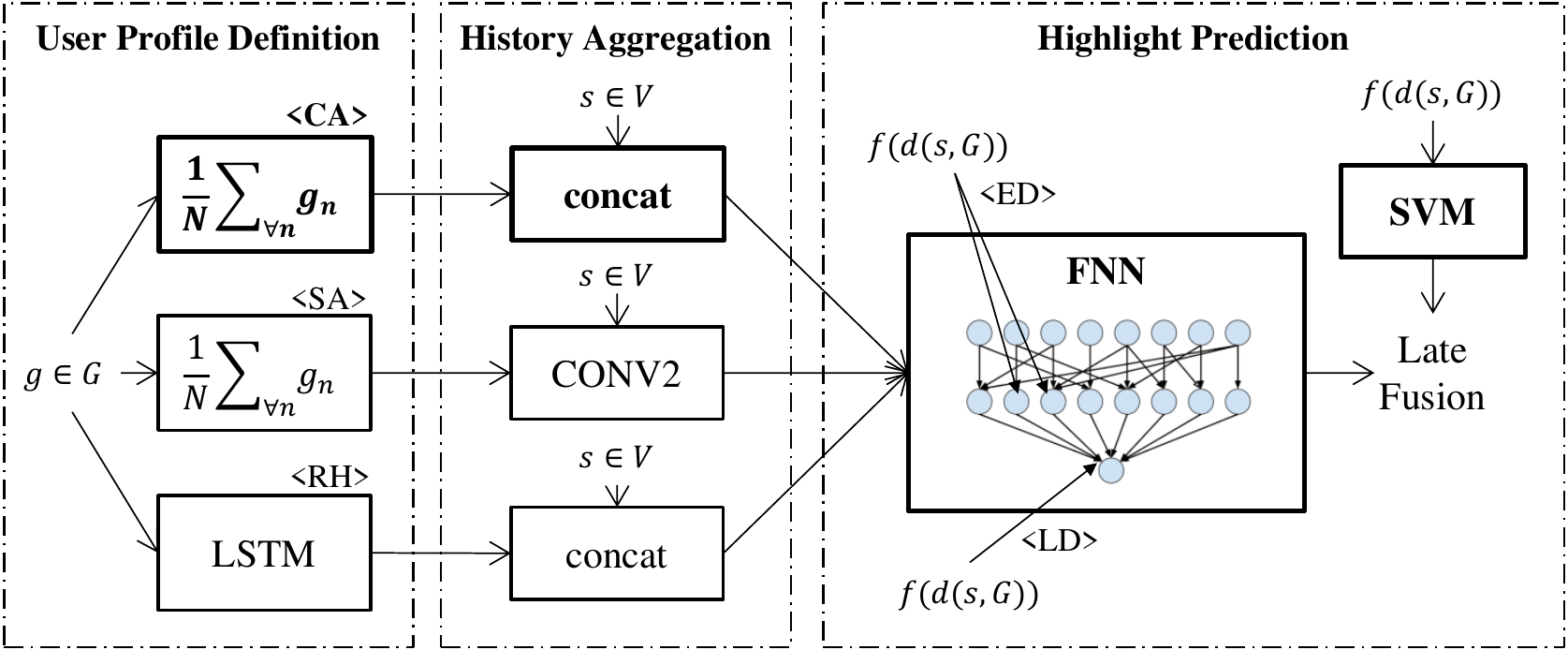}
    \caption{\textbf{Model architectures.} We show our proposed model (\textbf{bold}) and
    alternative ways to encode the history and fuse predictions (see section~\ref{sec:detail}).}
   \label{fig:method}
\end{figure}


\section{Experiments}
\label{sec:experiments}
We evaluate the proposed method, called \textbf{PHD-CA + SVM-D}, on the dataset introduced in Section~\ref{sec:dataset}. We start by comparing it against the state of the art for non-personalized highlight detection, as well as several personalization baselines  in Section~\ref{sec:stoa}.
Then, Section~\ref{sec:detail} analyses variations of our method and quantifies the contribution of the different inputs and architectural choices.

\paragraph{Evaluation metrics.}
We follow~\cite{gygli2016video2gif} and report mean Average Precision (mAP)
and normalized Meaningful Summary Duration, which rates how much of the video has to be watched before the majority of the ground truth selection was shown, if the shots in the video had been re-arranged to match the predicted ranking order. In addition, we report Recall@5, i.e. the ratio of frames from the user-generated GIFs (the ground truth) that are included in the $5$ highest ranked GIFs.


\begin{table}[t!]
\centering
\setlength\extrarowheight{1ex}
\setlength{\tabcolsep}{2.5pt}
\begin{tabular}[width=\textwidth]{@{}lr|HH@{}c@{ }H|HH@{}c@{ }H|HHHH@{}c@{ }H|P{3.5em}@{}}
& \textbf{Model}&\multicolumn{4}{@{}c@{}}{\textbf{mAP}}&\multicolumn{4}{@{}c@{}}{\textbf{nMSD}}&\multicolumn{2}{@{}c@{}}{
}&\multicolumn{4}{@{}c|}{\textbf{R@5}}&\textbf{Notes}\\
\hline \hline
& Random & 12.28\% & - & 12.97\% & - & 50.47\% & - & 50.60\% & - & 7.07\% & 7.06\% & 17.33\% & - & 21.38\% & - & \\
\hline
\multirow{3}{0.2cm}{\begin{sideways}{Non-personal \ }\end{sideways}} & 
Video2GIF \cite{gygli2016video2gif}&12.65\% & - & 15.69\% & - & 52.80\% & - & 42.59\% && -  6.76\% & 8.24\% & 17.45\% & -  & 27.28\% & - & Trained on~\cite{gygli2016video2gif} \\ 
& Highlight SVM & 13.96\%&- & 14.47\%&- & 47.14\%&- & 45.55\% &-& 8.14\% & 8.00\% & 22.04\% &-& 26.13\% &-& \\
& Video2GIF (ours) & 15.30\%& \textbf{15.67\%} & \textbf{15.86\%} & \textbf{16.15\%}  & 43.66\% & \textbf{42.92\%} & \textbf{42.06\%}& \textbf{41.26\%}  & 8.38\% & 8.39\% & 23.63\% & 24.92\% & \textbf{28.42\%} & \textbf{28.92\%}& \\
\hline \hline 
\multirow{4}{0.2cm}{\begin{sideways}{Personal}\end{sideways}} 
& Max Similarity & 15.43\%&- & 15.49\% &-& 44.12\% &-& 44.22\% &-& 8.17\% & 8.02\% & 27.40\% &-& 26.44\% &-& unsup.\\
& V-MMR & 14.36\% && 14.86\% && 44.82\% && 43.72\% && 8.28\% && 28.84\% && \textbf{28.22\%} && unsup.\\

& Residual & 14.89\% && 14.89\% && 47.07\% && 47.07\% && 7.94\% && 26.05\% && 26.05\% &&\\
& SVM-D & 15.23\% &-& \textbf{15.64\%} &-& 44.15\% &-& \textbf{43.49\%} &-& 8.34\% & 8.45\% & 27.77\% &-& 28.01\% &-& \\
\cline{2-17} 
\multicolumn{2}{@{}r|}{Ours (CA + SVM-D)} & 16.68\% && \textbf{16.68\%} && 40.58\% && \textbf{40.26\%} && 9.20\% && 31.16\% && \textbf{30.71\%} && \\
\end{tabular}
\caption{State-of-the-art comparison (videos segmented into $5$-second long shots). For mAP and R@5, the higher the score, the better the method. For MSD, the smaller is better. Best result per category in \textbf{bold}.}
\label{tab:stoa}
\end{table}

\subsection{Baseline comparison}
\label{sec:stoa}
We compare our method against several strong baselines:

\textbf{Video2GIF~\cite{gygli2016video2gif}.} This work is the state of the art for automatic highlight detection for GIF creation. We evaluate the pre-trained model which is publicly available. As the model is trained on a different dataset we additionally provide results for a slight variation of~\cite{gygli2016video2gif}, trained on our dataset, which we refer to as \textit{Video2GIF (ours)}.

\textbf{Highlight SVM.} This model is a ranking SVM~\cite{lee2014large} trained to correctly rank positive and negative segments as per Eq.~\eqref{eq:objective}, but only using the segment's descriptor and ignoring the user history.

\textbf{Maximal similarity.} This baseline scores segments according to their maximum similarity with the elements in the user history $\mathcal{G}$. We use the cosine similarity as a similarity measure.

\textbf{Video-MMR.} Following the approach presented in \cite{li2010multi}, $\mathcal{G}$ is used as query so that the segments that are most similar are scored highly. Specifically, we use the mean cosine similarity to the history elements $g_i$ as an estimate of the relevance of a segment.

\textbf{Residual Model.} Inspired by~\cite{ren2017personalized}, we include a residual model for ranking. 
\cite{ren2017personalized} proposes a generic regression model and a second user-specific model that personalizes predictions by fitting the residual error of the generic model.
To adapt this idea to the ranking setting, we propose training a user-specific ranking SVM that gets the generic predictions from \textit{Video2GIF (ours)} as an input, in addition to the segment representation $\mathbf{s}$.
Thus, a user's model is defined as

\begin{equation}
\label{eq:res}
h_{res}(s, \mathcal{G}) = \mathbf{w_\mathcal{G}}^{\mathbf{T}}\begin{bmatrix}
           \mathbf{s} \\
           h_{V2G}(\mathbf{s})
         \end{bmatrix} + b,
\end{equation}
where $\mathbf{w_\mathcal{G}}$ are the weights learned from the history $\mathcal{G}$.

\textbf{Ranking SVM on the distances.} This model corresponds to the model presented in Section~\ref{sec:distance-svm}.

\begin{figure}[t]
	\centering
	\begin{subfigure}[b]{1\linewidth}
		\centering 	\includegraphics[width=1\linewidth]{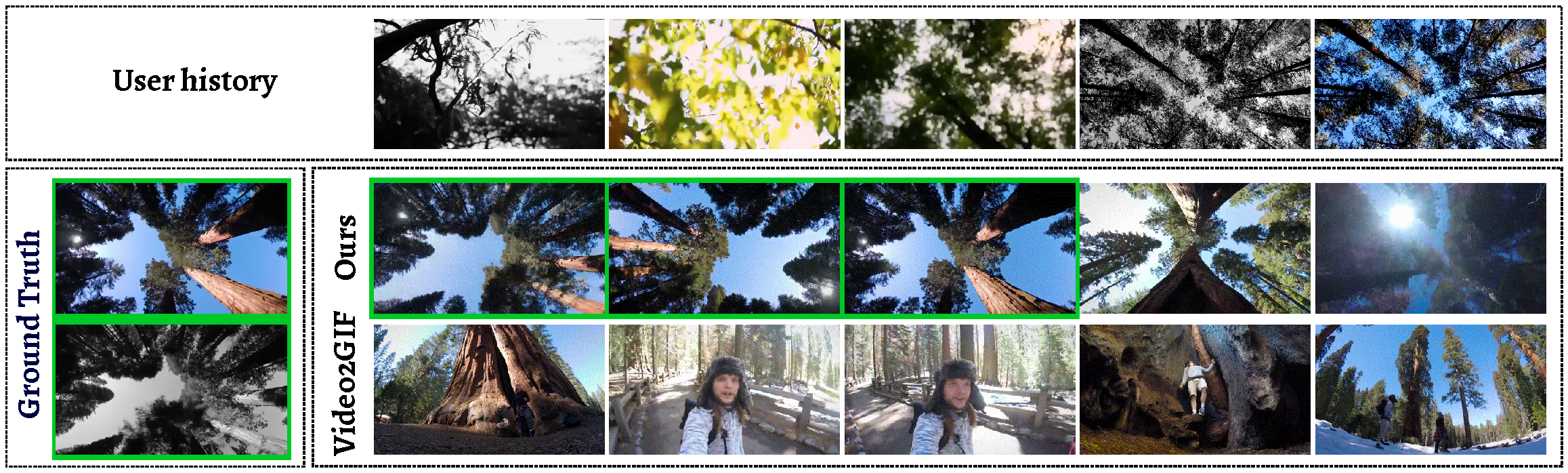}
		\caption{User with interest in forests}
		\label{fig:ex1}
	\end{subfigure}
	\begin{subfigure}[b]{1\linewidth}
		\centering 	\includegraphics[width=1\linewidth]{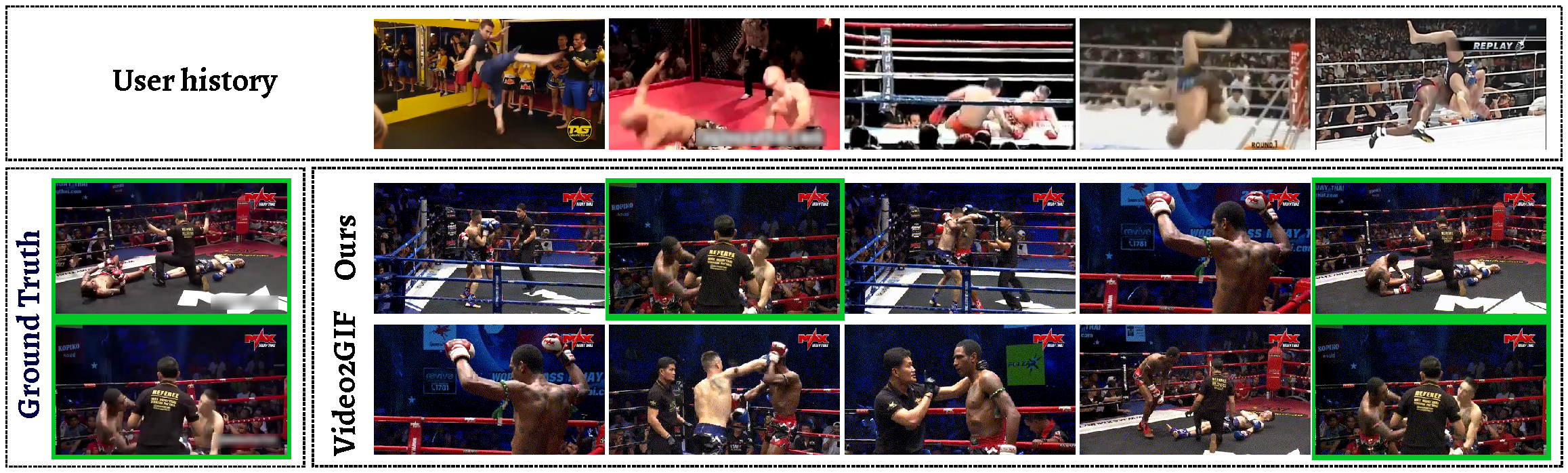}
		\caption{User favouring knock-outs}
		\label{fig:ex2}
	\end{subfigure}
	\begin{subfigure}[b]{1\linewidth}
		\centering 	\includegraphics[width=1\linewidth]{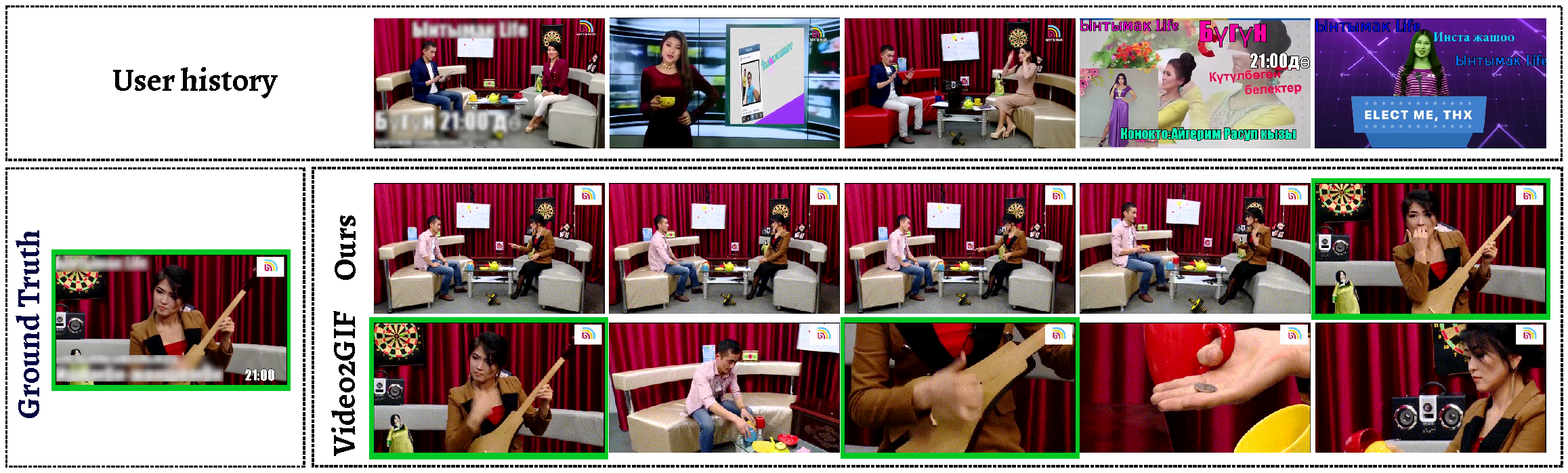}
		\caption{User creating previews for TV shows.}
		\label{fig:ex3}
	\end{subfigure}
	\caption{\textbf{Qualitative Examples.} We compare our method (\textbf{PHD-CA + SVM-D}) to generic highlight detection (\textbf{Video2GIF (ours)}). Videos for which personalization improves the Top 5 results are shown in \textbf{(a)} and \textbf{(b)}. In both cases the users are consistent in what content they create GIFs from. Thus, personalization provides more accurate results (correct results have \textcolor{darkgreen}{green borders}). In \textbf{(c)} we show a failure case, where the user history is misleading the model.}
	\label{fig:qualitative_results}
\end{figure}

\paragraph{Results.}
We show quantitative results in Table~\ref{tab:stoa} and qualitative examples in Figure~\ref{fig:qualitative_results}. When analyzing the results, we find that our method outperforms~\cite{gygli2016video2gif} as well as all baselines by a significant margin.
Adding information about the \textit{user history} to the highlight detection model (\textbf{Ours (CA + SVM-D)}) leads to a relative improvement over generic highlight detection (\textbf{Video2GIF (ours)}) of \att{5.2\% (+0.8\%) in mAP, 4.3\% (-1.8\%) in mMSD and 8\% (+2.3\%) in Recall@5}.
This is a significant improvement in this challenging high-level task and compares favorably to
\att{the improvement obtained in}
previous work~\cite{gygli2016video2gif}.
The improvement of our method over using the user history alone is even larger, thus reinforcing the need to train a personalized highlight detection model that uses the information about all users jointly.

Models using only generic highlight information or only the similarity to previous GIFs perform similar (15.86\% for \textbf{Video2GIF (ours)}  \textit{vs.} 15.64\% mAP for \textbf{SVM-D}), despite the simplicity of the distance model. Thus, we can conclude that these two kind of information are both important and that there is a lot of signal contained in a user's history about his future choice of highlights. 
\att{
This concurs with our qualitative analysis in Section~\ref{sec:data_analysis}, where we find that
that most users in our dataset show high consistency in the kind of highlights they selected.}

Given that the combination of the two kinds of information improves the final results, we conclude that they are complementary to each other and that it is beneficial to use models that consider them both. The residual model also combines generic highlight detection and personalization. It however estimates model weights per user, which leads to inferior results on our dataset, due to the small number of training examples per user. Indeed, the \textbf{Residual} baseline is outperformed by the generic highlight detection and the personalization baselines, in particular \textbf{SVM-D}. Our method, on the other hand, performs well in this challenging setting and outperforms all baselines by a large margin.

To better understand how the model works, Figure~\ref{fig:qualitative_results} shows qualitative results for our method and a non-personalized baseline, along with the user history. 
As can be seen from \ref{fig:ex1} \& \ref{fig:ex2}, our method effectively uses the history to make more accurate predictions.
In~\ref{fig:ex3} we show a failure case, where the history is not indicative of the highlight chosen by the user.

\subsection{Detailed experiments}
\label{sec:detail}

\begin{figure*}[t]
	\centering
	\includegraphics[width=.6\linewidth]{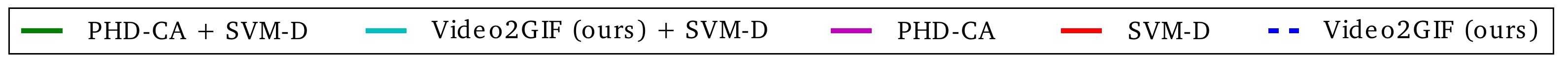}
	\vspace{1ex}
	
	\begin{subfigure}[b]{0.33\linewidth}
		\centering 	\includegraphics[width=1\linewidth]{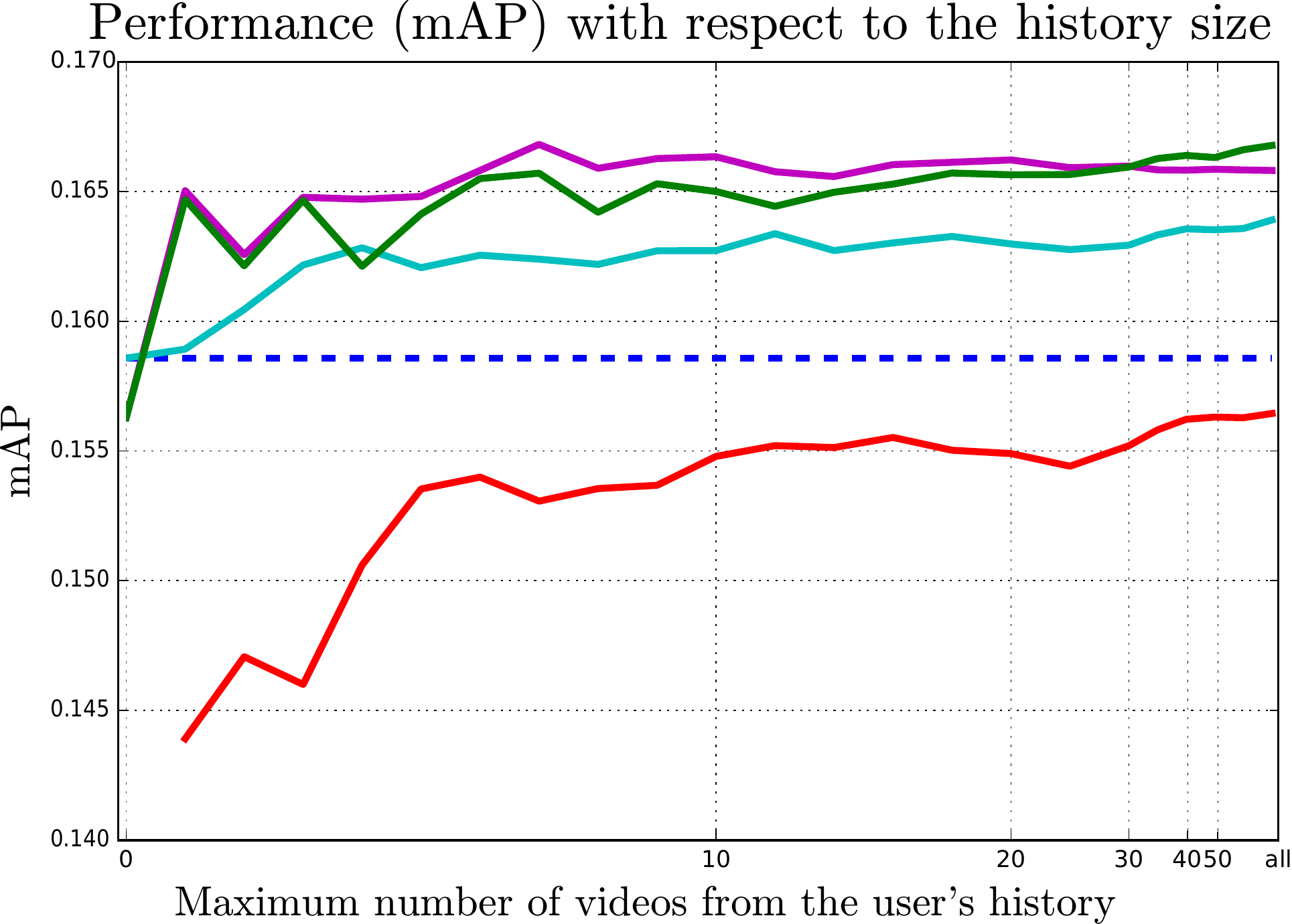}
		\caption{\label{fig:history_vs_map}}
	\end{subfigure}
	\begin{subfigure}[b]{0.33\linewidth}
		\centering 	\includegraphics[width=1\linewidth]{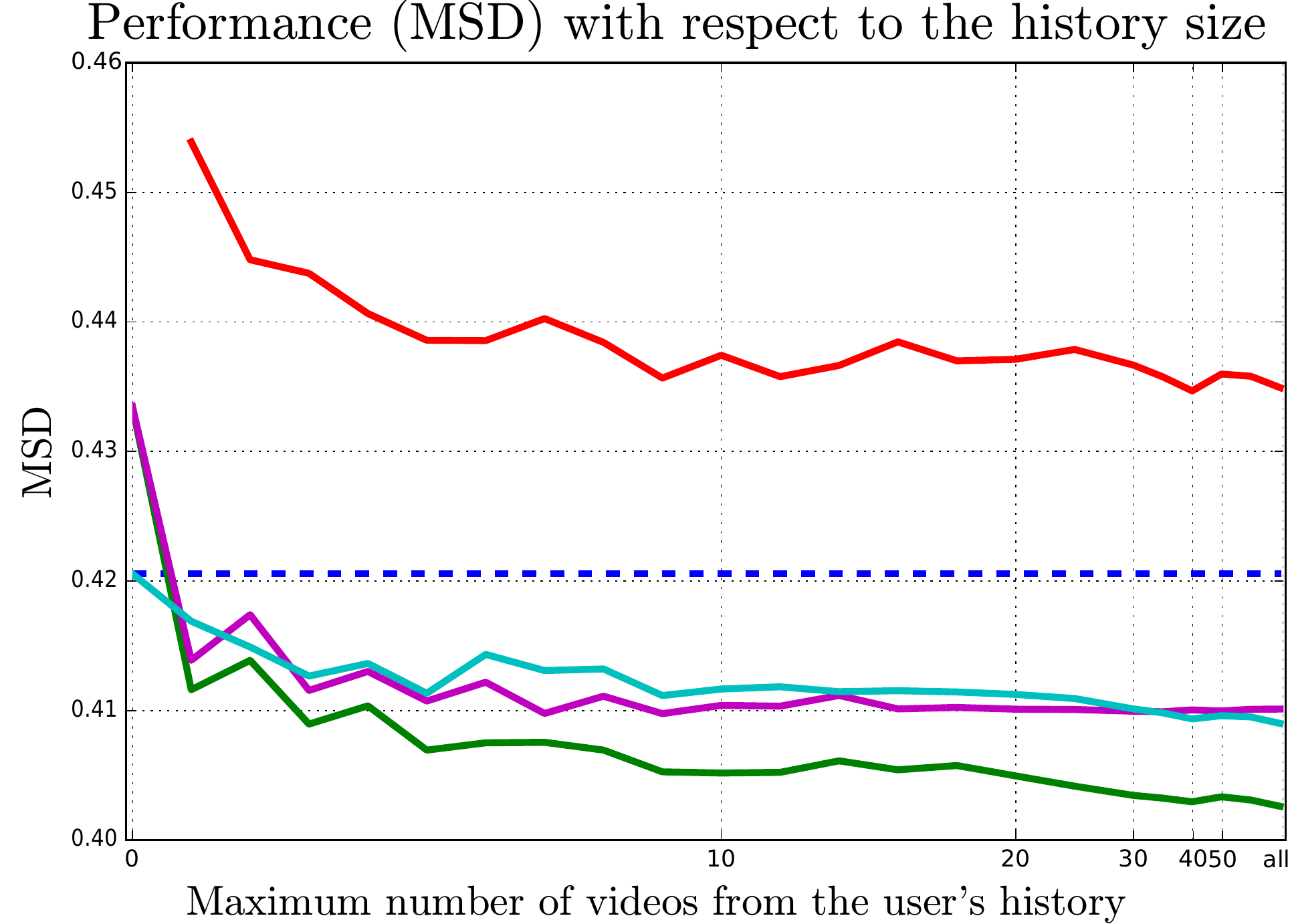}
		\caption{\label{fig:history_vs_msd}}
	\end{subfigure}
	\begin{subfigure}[b]{0.33\linewidth}
		\centering 	\includegraphics[width=1\linewidth]{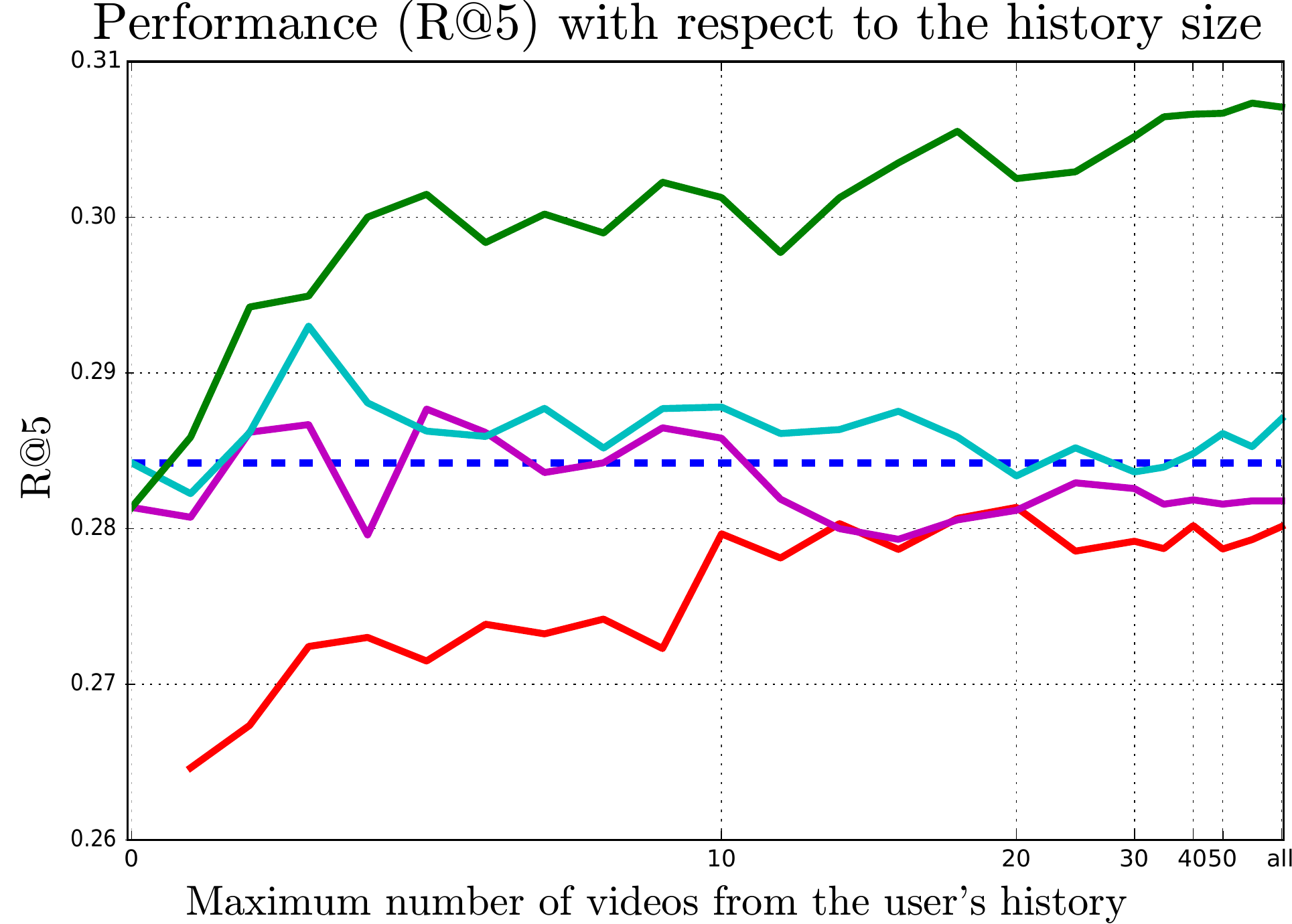}
		\caption{\label{fig:history_vs_recall}}
	\end{subfigure}
	\caption{Performance of different methods as a function of the history size. We observe that our method improves over generic highlight detection with as little as one history element per user.
		Furthermore, performance has not saturated even when using the full history, thus indicating that our method can effectively use longer histories as well.
		Interestingly, we find that only models including the distances to the history as a feature improve Recall@5,~\ie provide better results at the top of the ranking. 
		\textbf{Best viewed in color.}}
	\label{fig:history_vs_performance}
\end{figure*}

In the following, we analyze different variations of our approach.
In particular, we compare various ways to include the user history, network architectures, and fusion of different inputs. Figure~\ref{fig:method} shows these different configurations, while their performance  is given in Table~\ref{tab:results_detail}.
Additionally, we analyze the performance of our model as the size of the user history varies (Figure~\ref{fig:history_vs_performance}).

\paragraph{Learning an aggregation \emph{vs} averaging?}
Our proposed model aggregates the history via averaging (\textbf{PHD-CA}, \cf Section~\ref{sec:FNN}). Alternatively,
\att{Recurrent Neural Networks are often successfully used to encode visual sequences~\cite{srivastava2015unsupervised, GarciadelMolino2018predicting}. Thus, we also explored a model that uses an LSTM to learn to aggregate the history (\textbf{PHD-RH}).}
The history is then concatenated to the segment representation and passed through 2 fully-connected layers.
As can be seen from Table~\ref{tab:results_detail}, having a predefined aggregation performs better than learning it. We attribute this to the challenge of learning a sequence embedding from limited data and 
\att{conclude that an average aggregation provides an effective representation of the users' history}.

\paragraph{Convolutional combination or concatenation?}
In Section~\ref{sec:FNN} we propose to concatenate the average history to the segment representation $\mathbf{s}$. Since they both use the same C3D representation, however, it is also possible to first aggregate each dimension of the two vectors with 1D convolutions, before passing them through fully connected layers (\textbf{PHD-SA}). We compared these two approaches and found the concatenation to give superior performance.
The convolutional aggregation uses the structure of the data to reduce the number of network parameters and therefore has roughly half the parameters of the concatenation model. Convolutional aggregation, however, requires the network to aggregate the history into the segment information per dimension, using the same weights. Thus it is limited in its modeling capacity, compared to a network using concatenated features as inputs.

\iffalse

\begin{table}[t!]
\centering
\setlength\extrarowheight{1ex}
\setlength{\tabcolsep}{2.5pt}
\begin{tabular}[width=\textwidth]{@{}r|HHcH|HHcH|HHHHc@{}HH}
\textbf{Model}&\multicolumn{4}{c}{\textbf{mAP}}&\multicolumn{4}{c}{\textbf{nMSD}}&\multicolumn{2}{c}{
}&\multicolumn{4}{@{}c}{\textbf{R@5}}&\\
\hline \hline
Video2GIF (ours) & 14.53\% && 14.53\% && 43.73\% && 43.73\% && 8.24\% && 26.03\% && 26.03\% &&
\\
Max Similarity & 14.49\% && 14.49\% && 42.78\% && 42.78\% && 8.22\% && 23.84\% && 23.84\% &&
\\
SVM-D& 14.91\% && 14.91\% && 41.94\% && 41.94\% && 8.52\% && 25.32\% && 25.32\% &&
\\
PHD-RH & 14.23\% && 14.23\% && 44.62\% && 44.62\% && 7.96\% && 25.15\% && 25.15\% &&
\\
PHD-SA& 14.36\% && 14.36\% && 42.97\% && 42.97\% && 8.05\% && 25.75\% && 25.75\% &&
\\
PHD-CA& 15.17\% && 15.17\% && 41.54\% && 41.54\% && 8.48\% && 26.89\% && 26.89\% &&
\\
\hline
PHD-CA-ED (1st layer)& 14.35\% && 14.35\% && 42.19\% && 42.19\% && 7.83\% && 24.31\% && 24.31\% &&
\\
PHD-CA-LD (last layer)& 14.66\% && 14.66\% && 41.20\% && 41.20\% && 8.22\% && 26.51\% && 26.51\% &&
\\
Video2GIF(ours) + SVM-D& 15.37\% && 15.37\% && 41.75\% && 41.75\% && 8.76\% && 27.73\% && 27.73\% &&
\\
PHD-CA + SVM-D& 15.54\% && 15.54\% && 39.38\% && 39.38\% && 8.35\% && 27.51\% && 27.51\% &&
\\

\end{tabular}
\caption{Detailed experiments on the validation set. We analyze different ways to represent and aggregate the history, as well as ways to use the distances to the history to improve the prediction.}
\label{tab:results_detail}
\end{table}

\else
\begin{table}[t!]
\centering
\setlength\extrarowheight{1ex}
\setlength{\tabcolsep}{2.5pt}
\begin{tabular}[width=\textwidth]{@{}r|HHcH|HHcH|HHHHc@{}HH}
\textbf{Model}&\multicolumn{4}{c}{\textbf{mAP}}&\multicolumn{4}{c}{\textbf{nMSD}}&\multicolumn{2}{c}{
}&\multicolumn{4}{@{}c}{\textbf{R@5}}&\\
\hline \hline
PHD-SA & 14.93\% && 15.73\% && 43.76\% && 42.80\% && 8.52\% && 24.24\% && \textbf{28.65\%} &&\\
PHD-RH & 14.79\% & 15.62\% & 15.74\% & 16.06\% & 43.98\% & 42.35\% & 42.75\% & 41.68\% & 8.28\% && 22.32\% & 28.83\% & 27.45\% & 29.91\% &\\
PHD-CA & 15.37\% && \textbf{16.58\%} && 42.67\% && \textbf{41.01\%} && 8.50\% && 24.72\% && 28.18\% &&\\
\hline
PHD-CA-ED (1st layer) & 15.30\% && 16.14\% && 43.04\% && 41.26\% && 8.70\% && 24.46\% && 29.20\% &&\\
PHD-CA-LD (last layer) & 15.71\% && 16.20\% && 42.84\% && 41.07\% && 8.76\% && 25.42\% && 29.78\% &&\\
Video2GIF (ours) + SVM-D & 16.68\% && 16.39\% && 40.81\% && 40.90\% && 9.01\% && 30.53\% && 28.70\% && \\ 
\textbf{PHD-CA + SVM-D} & 16.68\% && \textbf{16.68\%} && 40.58\% && \textbf{40.26\%} && 9.20\% && 31.16\% && \textbf{30.71\%} &&\\
\end{tabular}
\caption{Detailed experiments. We analyze different ways to represent and aggregate the history, as well as ways to use the distances to the history to improve the prediction.}
\label{tab:results_detail}
\end{table}
\fi

\paragraph{Adding distances, with early or late fusion?}
As we discussed, our assumption is that the similarity of a segment to the previously chosen GIFs is informative when predicting the score of a segment. Thus, we tested models that use the distance to the history elements as an additional input.

Since using distance features leads to a different representation compared to the feature activations of C3D, it is unclear how to best merge the two different modalities. We tried early fusion (concatenation of the two inputs, \textbf{PHD-CA-ED}), late fusion before the prediction layer in one single model (\textbf{PHD-CA-LD}) and late fusion with training two separate models (\textbf{PHD-CA + SVM-D}), as shown in Figure~\ref{fig:method}. We find that late fusion performs superior to early fusion, and that combining two different models outperforms merging on the last layer of the neural network.
The superiority of late fusion is to be expected, as neural networks often struggle to combine information from different modalities~\cite{simonyan2014two, carreira2017quo}.
Adding the distances in the neural network even slightly decreases mAP, while Recall@5 improves. 
While this inconsistency is somewhat surprising, Recall@5 is arguably more important, as it evaluates the accuracy of the top-ranked elements, which is what matters for finding highlight in videos, while mAP considers the complete ranking.
When using a separate model for the distances and fusing their predictions, we obtain a consistent improvement in all metrics.

We also tried adding personalization to a generic highlight detection model by combining its predictions with the predictions of the distance SVM (\textbf{Video2GIF (ours) + SVM-D} in Table~\ref{tab:results_detail}). This leads to a significant improvement over the generic model. While it doesn't perform quite as well as our full model, this approach provides a simple way to personalize existing highlight detection, in order to improve their performance.

\paragraph{How much does personalization help for different history sizes?}
We are interested in how well the model performs when very little user-specific information is available.
To do so, we restrict the history provided to the model to the last $k$ videos a user created GIFs from, rather than providing the full history\footnote{Note that some users may have less than $k$ videos in their history, and only $n<k$ videos can be considered.}.

We plot the performance as a function of the history length $k$ in Figure~\ref{fig:history_vs_performance}.
From this plot, we make several important observations.
(i) Adding personalization helps even for small histories.
Recall@5 improves by \att{5.6\% (+1.6\%)} over the generic model for a history size of $k=4$, for example.
Even for $k=1$,~\ie a single history video, our method outperforms generic highlight detection across all metrics. Having a model that performs well  given few history elements is important, as the history size in our dataset follows a long tail distribution (\cf Figure~\ref{fig:DBstats}).
Indeed, we discarded more than 90\% of the user profiles when creating our dataset, as they had a history of fewer than 5 elements.
(ii) While \textbf{PHD-CA} quickly improves mAP as the history grows, only the model including the distances significantly improves Recall@5. This is consistent with our experiments in Table~\ref{tab:results_detail}.
Improving the ranking of the highest scoring segments is challenging, as they often have only subtle differences.
The similarity to a user's history allows to capture these differences and thus obtain a better ordering of the top elements.
(iii) Performance is not yet saturated for the history lengths in the dataset. Thus our model is not only able to make use of small histories, but can also effectively use larger histories to further improve prediction accuracy.

\subsection{Implementation details}
\label{sec:implementation}
\paragraph{Data Setup:} 

The dataset consists of a total of $13,822$ users, of which $11,972$ are used for training, $1,000$ for validation, and $850$ for testing.
At both training and test time, the goal of our models is to predict what part of video $V$ a user chooses, given his history $\mathcal{G}$.
As such, $V$ corresponds to the last video from each user, and all other videos are used to build each user's history $\mathcal{G}$. The validation set is used to find the best hyper-parameters for the highlight models and also to find the right weight $\omega$ for Eq.~\ref{eq:LF}.

\begin{figure}[t]
	\centering
	\includegraphics[width=.9\linewidth]{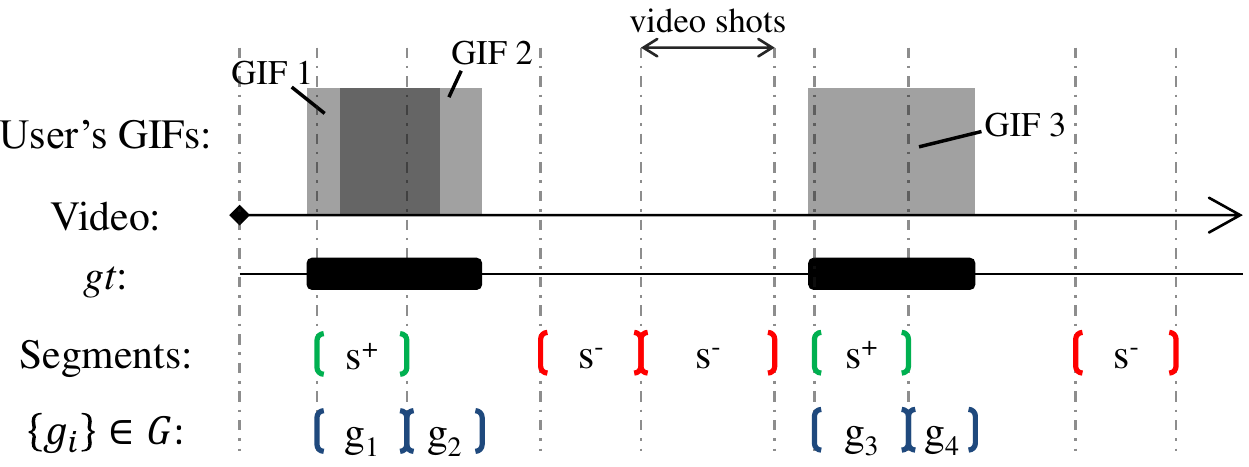}
	\vspace*{-2pt}
	\caption{Procedure to obtain the pairs of segments $\left( s^+,s^- \right)$ in $V$, the user selection $gt$ for the evaluation, and the user \emph{history} $g \in \mathcal{G}$ from any other $video\neq V$.}
    \label{fig:sampling}
\end{figure}

To train our models, we have sampled five positive-negative pairs \(\left( s^+,s^- \right)\) from each user's video $V$, where a positive example $s^+$ is a shot that was part of the user's GIFs for that video (see Figure~\ref{fig:sampling}), and a negative example $s^-$ is a shot that was not included in any GIF.
To split the user selected segments into shots, we use the shot detection of~\cite{Gygli17DeepCut} and deterministically split shots longer than $15$ seconds into $5$ second chunks.
For the user history $\mathcal{G}$, we use a maximum of $20$ shots, which are selected at random ensuring that there is at least one shot from each of the last $k=20$ videos in the user's history.
Since a user may generate several overlapping GIFs before being satisfied with the result, $\mathcal{G}$ (and analogously the ground truth for $V$) does not correspond to each of the user-generated GIFs independently, but rather their union.

At test time the videos are segmented into fixed segments of 5 seconds to be able to compare to~\cite{gygli2016video2gif}. Furthermore,~\cite{Gygli17DeepCut} may predict short shots and gaps (due to slow scene transitions), which, when used at test time, would lead to noise in the evaluation.
We use the user's full history when making predictions. Since the distance-based models require a $k$-dimensional input, the distance vector is filled with zeros if \att{$|\mathcal{G}| < k$, and the elements further away are discarded if $|\mathcal{G}| > k$}. 

\paragraph{Training methodology:} 

We optimize the network parameters using grid search over different possible FNN architectures. Different dropout values (random search between $.5$ and $.8$ for the input layer, and $.1$ to $.5$ for the intermediate ones) and activation functions (\emph{ReLU} and \emph{SELU}~\cite{klambauer2017self}) were explored, as well as the use of batch normalization~\cite{ioffe2015batch} after each layer. Using \textit{RMSProp} as optimizer and a weight decay between $1\mathrm{e}{-3}$ and $2$, the initial learning rate (randomly set between $1\mathrm{e}{-2}$ and $1\mathrm{e}{-4}$) is decreased by half every four epochs, for a total of $16$ epochs per search iteration. The pairwise loss function used for all models is $l_1$. Our models are implemented in TensorFlow~\cite{abadi2016tensorflow}.

For the aggregation of \(s \in V\) and \(g \in \mathcal{G}\), a size of either four or ten neurons is considered for the 1-D convolution in \textbf{PHD-SA}, flattened with a single neuron convolution before the FNN layers.
For the \textbf{PHD-RH} model, we tested using $1000$ or $512$ neurons in the hidden layer of the LSTM.

For learning the combination of the user profile and the segment information, we ran hyper-parameter search and varied the number of hidden layers of the FNN from 1 to 3.
We tested layers having up to $512$ neurons, where each following layer would have the same number or fewer neurons.
We find that smaller architectures perform best: Two hidden layers of $512$ and $64$ neurons for \textbf{PHD-CA} (with dropout of $.71$ and $.18$ in the input and intermediate layers, respectively); a single hidden layer of $256$ neurons for \textbf{PHD-SA}
; and a single hidden layer of $512$ neurons for \textbf{PHD-RH}. 


\section{Conclusion}
In this work, we proposed an approach for personalized highlight detection in videos.
The core idea of our approach is to use a model that is trained for all users jointly and which is customized via its inputs, by providing a user's previously chosen highlights at test time. Such an approach allows training a high-capacity model, even when few examples per user are available.
In our experiments, we have shown that the user history provides a
useful
signal for future selections and that incorporating that information into our highlight detection model significantly improves performance:
Our method outperforms generic highlight detection by \att{8\%} in Recall@5.
When training a separate model per user, as done in previous work, personalization does not outperform generic highlight detection. Our method, on the other hand, works well, even when given very few user-specific training examples. It outperforms generic highlight detection given just a single user-specific training example, thus confirming the benefit of our model architecture.

Finally, in order to train and test our model, we have introduced a large-scale dataset with user-specific highlights. To the best of our knowledge, \att{it is the first personalized highlight dataset at that scale and the first which is made publicly available.}

\bibliographystyle{ACM-Reference-Format}
\bibliography{biblio}

\end{document}